# How Different Text-Preprocessing Techniques using the BERT Model Affect the Gender Profiling of Authors

Esam Alzahrani[1, 2] and Leon Jololian[1]

[1]Department of Electrical and Computer Engineering,
University of Alabama at Birmingham, Birmingham, AL, USA
[2]Department of Computer Engineering, Al-Baha University,
Alaqiq, Saudi Arabia

## ABSTRACT

*Forensic author profiling plays an important role in indicating possible profiles for suspects. Among the many automated solutions recently proposed for author profiling, transfer learning outperforms many other state-of-the-art techniques in natural language processing. Nevertheless, the sophisticated technique has yet to be fully exploited for author profiling. At the same time, whereas current methods of author profiling, all largely based on features engineering, have spawned significant variation in each model used, transfer learning usually requires a preprocessed text to be fed into the model. We reviewed multiple references in the literature and determined the most common preprocessing techniques associated with authors' genders profiling. Considering the variations in potential preprocessing techniques, we conducted an experimental study that involved applying five such techniques to measure each technique's effect while using the BERT model, chosen for being one of the most-used stock pretrained models. We used the Hugging face transformer library to implement the code for each preprocessing case. In our five experiments, we found that BERT achieves the best accuracy in predicting the gender of the author when no preprocessing technique is applied. Our best case achieved 86.67% accuracy in predicting the gender of authors.*

## KEYWORDS

*Authorship profiling, NLP, digital forensics, transfer learning*

## 1. INTRODUCTION

Forensic author profiling has proven to be an important yet complicated task that requires further investigation. According to Keretna et al. [1], writing styles are affected by factors like a person's culture, educational background, and the environment he/she has been raised in. Historically, Criminal Forensic profiling is established by a former Federal Bureau of Investigation (FBI), John Edward Douglas [2]. He started studying and analyzing serial killers' crimes. Moreover, he has interviewed some of the serial killers to find a pattern that is associated with their criminal activities and their characteristics profiling. Further, this approach is believed to help to direct the course of investigation towards the most possible suspect who committed the crime. As the use of the Internet grows, the need to adopt this change is inevitable. The content of the internet contains a big amount of unstructured text especially in Online Social Networks (OSNs). One of the tasks of analyzing textual content is authorship profiling by which demographic information about anonymous authors can be revealed. Beyond that, authorship profiling can contribute to the deanonymization of anonymous malicious texts. Even though most OSNs are regarded as an auxiliary in which users use their real names, the option of anonymity is still available. Criminals are known to seek anonymity to avoid getting caught by law enforcement [3], [4]. Moreover, the considered attributes that are used to create similarity among different profiles, can be used to build profile clusters by which users' profiles can be assigned.

However, other gaps need to be addressed as well, including that no method of classifying age or gender—both of which are aspects of author profiling—that also considers the genre or nature of the analyzed text has been widely endorsed. Beyond that, to the best of our knowledge, no research has involved surveying or comparing the different approaches used in author profiling. In fact, current knowledge about the task is largely based on small-scale experiments conducted to find a reliable classification method with a near-zero error rate. In order to use forensic authorship profiling as admissible evidence in courts, the proposed methods must have about 100% accuracy. The average accuracy of good proposed methods are ranging from 70-85% [5]–[9]. However, the proposed methods still suffer from low accuracy. A more specific gap is that transfer learning, an emerging technique in natural language processing (NLP) proven to be state-of-the-art for many NLP tasks [10], has not been fully tested in the context of author profiling. As a consequence, literature on author profiling using transfer learning has remained slight, and how the technique contributes to and affects such profiling remains poorly understood, at least in a systematic sense. For those reasons, the applicability of the advanced, reliable technique of transfer learning to author profiling is worth investigating.

According to Devlin and Chang, a serious challenge in NLP is the limited amount of training data [11]. To address that challenge, researchers have developed transformer-based models that are trained on enormous unlabelled datasets—for instance, Wikipedia's dataset—so that researchers can use the pretrained models on smaller datasets instead of having to develop training models from scratch. Although the technique has been shown to afford more accuracy in executing different NLP tasks [11], [12], the pretrained models require so-called "fine-tuning" before being used with smaller datasets. An example of such a pretrained model is the bidirectional encoder representations from transformers (BERT) model, which is distinguished from other pretrained models by virtue of its bidirectionality—that is, it considers context when words have dual valence. For example, when processing the word bank, which can mean a financial institution or the shore of a river, the BERT model examines all words in the sentence at both valences and generates a score that indicates the best representation of the meaning of the words in their given context. In its implementation, the BERT model is based on transformer model architecture developed by researchers at Google in 2017 [11].

The purpose of this study is to examine the impact of the most used preprocessing techniques in profiling the age and gender of the author if a pretrained model is used, BERT. In the remainder of this paper, the next section introduces past work related to the preprocessing techniques in author profiling, followed by an experimental section that details the implementation of the five cases of preprocessing techniques that we considered, and the steps performed in each experiment. After that, a section presents the results of each experiment and discusses the effect of each preprocessing technique on the model's accuracy in predicting the gender of authors. In the paper's conclusion, we restate the important findings of our study and indicate directions for future work.

## 2. RELATED WORK

To conduct a thorough literature review, we considered the valuable contribution of PAN's shared tasks in author profiling, which provides a broad range of approaches and methods in the field. We reviewed PAN's shared tasks from 2013 to 2017 to identify the types of preprocessing techniques that researchers consider in their efforts to profile authors [13]–[17]. Besides, we also included papers that investigate author profiling using English datasets. Some papers used uncommon preprocessing techniques e.g., extending shortened texts such as slang words, contractions, and abbreviations [18]. Lundeqvist & Svensson removed HTML tags, and used Twitter custom tokenizer (nltk.tokenize package — NLTK 3.6.2 documentation)[19]. However, some papers considered common preprocessing techniques, similar to those used in PAN shared tasks [20]. At least five research groups represented in PAN's shared tasks from 2013 to 2017 removed retweet tags from the texts during preprocessing [13]–[17]; 17 groups removed hashtags [13]–[17], [20]; and 19 teams considered removing URLs. For the removal of the mentioned tags, 17research groups considered removing them from the processed text. Stop words were removed only four times [21]–[23] [24], and 29 teams did not apply any preprocessing technique whatsoever [13]–[17]. In 11 instances, retweet tags, URLs, and mentions were all removed [14]–[17], [25].

The effectiveness of preprocessing techniques in machine learning approaches depends on the selection of features and classifiers. In transfer learning, the extensive training of pretrained models on large data equips them with the needed power to model the language and capture most of its contextualized aspects. Because transfer learning uses the previously learned knowledge to tokenize the downstream text [26], it uses fine-tuning to train and classify the downstream task [27].

Some techniques in NLP, including transfer learning, have been proven to be state-of-the-art for many NLP tasks [10], however, not fully tested for author profiling. A systematic understanding of how transfer learning contributes to author profiling is also lacking, despite the clear value of investigating the applicability of such an advanced, reliable technique in author profiling. Although transfer learning is a technique with growing interest, publications on author profiling using it have been few. With this publication, we aim to narrow all of those gaps, at least in part.

## 3. EXPERIMENTS

### 3.1. Dataset

Table 1. The distribution of the dataset per class. The raw data was extracted from the URLs that were sent by PAN. The table illustrates the number of tweets per men and the number of tweets per women.

| | |
|---|---|
| **Number of authors** | 436 |
| **Number of tweets** | 363,031 |
| **Number of tweets per men** | 149,059 |
| **Number of tweets per women** | 113,972 |

The dataset we used is from PAN's 2016 shared tasks involving author profiling[16]. We chose to conduct our experiments on the English corpus only due to the focus and scope of our study. The most studied datasets in the literature are collected from Twitter. Twitter texts represent the characteristics of today's text e.g., unstructured, short, and colloquial. In PAN shared task in 2016, the participated research groups were sent the URLs of the tweets with a truth table containing the authors' gender and age labels. The URLs of the tweets have relatively smaller size and easier to share compared to the complete textual dataset. Age was categorized as follows: 1) 18–24, 2) 25–34, 3) 35–49, 4) 50–64, and 5) 65 and older. Table 1 shows the distribution of the dataset per class. Given the aim of our study, we profiled the author's gender only.

## 3.2. Experimental Setups

The adopted experimental setups were based on the most common techniques observed in the literature [13]–[17], all of which have been extensively tested in the context of author profiling using machine and deep learning techniques. The effect of preprocessing techniques on author profiling using transfer learning techniques has not yet been studied, however. As illustrated in Table 2, we considered five cases for the preprocessing techniques. In Case 1, we included three basic techniques: mentions removal, retweet tags removal, and hashtags removal. In Case 2, we added URLs removal to the techniques from Case 1, and in Case 3, we added the removal of punctuation. In Case 4, we applied a well-known technique in NLP, stop words removal, which involves eliminating extremely common words that are liable to be repeated in many texts and that some researchers characterize as noise, not as markers. Last, in Case 5, we chose not to apply any preprocessing technique in order to gauge its effect. We built all five cases using regex in Python and the Hugging Face transformer library on Google Colab.

Table 2. Preprocessing cases. The preprocessing techniques are explained for each case.

| Case | Preprocessing techniques |
| --- | --- |
| Case 1 | Mentions removal |
|  | Retweet tags removal |
|  | Hashtags removal |
| Case 2 | Mentions removal |
|  | Retweet tags removal |
|  | Hashtags removal |
|  | URLs removal |
| Case 3 | Mentions removal |
|  | Retweet tags removal |
|  | Hashtags removal |
|  | URLs removal |
|  | Punctuation removal |
| Case 4 | Mentions removal |
|  | Retweet tags removal |
|  | Hashtags removal |
|  | URLs removal |
|  | Punctuation removal |
|  | Stop words removal |
| Case 5 * | None (i.e., each text as-is) |

*No preprocessing technique was applied

All of the experiments were carried out using Google Colab's graphical processing unit (GPU) to optimize the time efficiency. We chose to run each case's code for three epochs, as suggested by the BERT model's developers [11], and we separated each code for each case and manually double-checked the effect of the preprocessing technique performed on the dataset. The rest of the code concerns the implementation of the BERT model for binary classification: 0 for authors who are men, 1 for authors who are women. The only difference in each experiment was in preprocessing; the rest of the experiment parameters were controlled and the same.

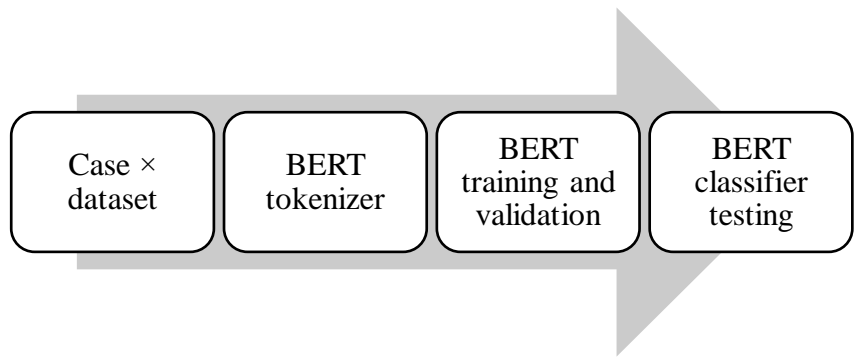

Figure 1. Experiments pipeline. The sequence of the conducted experiments. The case number changes based on the case we are considering. The rest of the experiment is the same for all cases.

Figure 1 illustrates the steps of our implementation which can summarized as follows:

1. Dataset handling and preprocess:
1.1. Reading the dataset and storing in dataframe
1.2. Preprocessing based on the target case x (x=1-5), see table 2
1.3. Loading BERT tokenizer
1.3. Building PyTorch dataset
1.4. Splitting Dataset into 90% training dataset and 10% testing dataset.
1.5. Building dataloader using BERT tokenizer, Batch_size = 32, and MAX_LENGTH = 100
2. Model fine-tuning:
2.1. Loading bert-base-cased from Huggiingface library
2.2. Building a binary classifier using BERT pretraining parameters with learning rate = 2-5
2.3. Implementing functions for training and evaluation
2.5 Iterating the training and evaluation for the number of epochs =3
3. Testing the model:
3.1. Implementing testing function using the trained classifier
3.2. Classifying testing dataset and compare prediction with real values
3.3. Creating a classification report for f1-score

The model that we used is the uncased BERT base model with hyperparameters listed in Table 3. For our study, the values of the hyperparameters and the type of the BERT model considered were not in focus; therefore, we did not dedicate much time to experimenting with different values of the hyperparameters. To better measure the effect of the various preprocessing techniques, we controlled the values of the hyperparameters in all five experiments.

Because transfer learning does not require much consideration of features engineering, the only obvious parameter that can vary from one study to another is the preprocessing techniques. The obvious advantage of examining those cases is to determine the best preprocessing technique for author profiling when transfer learning is used.

Table 3. Values of the hyperparameters.

| Parameter | Value |
|---|---|
| **Model** | BERT-base-cased |
| **Epochs** | 3 |
| **Batch size** | 32 |
| **Text max length** | 100 |
| **Learning rate** | $2^{-5}$ |

## 4. RESULTS AND DISCUSSION

In our study, we sought to examine the effect of the most commonly used preprocessing techniques on the gender profiling of authors when using a pretrained model: the BERT model. To distinguish our five cases of preprocessing techniques from each other, we conducted an experiment for each case. Because transfer learning models were trained on a relatively large dataset, we thought that the pretrained models would perform better in downstream tasks when the downstream dataset was larger. After performing the five experiments, we found that the best case for BERT was not applying any preprocessing technique and that the worst was Case 4, when we applied five preprocessing techniques (i.e., mentions removal, retweet tags removal, hashtags removal, URLs removal, and punctuation removal). Removing stop words also negatively affected the use of the BERT model. The rest of the cases differed slightly in accuracy, as shown in Table 4. The difference between Case 4 and Case 5 contributed to a significant difference overall (i.e., approx. 8%). The fewer preprocessing techniques we applied, the higher accuracy we observed. However, the time needed to train the model for Case 5 was the longest.

Table 4. Results of experiments. We consider using cross-validation to test the accuracy of the built models. We split the dataset into 90% for training and 10% for testing.

| Case | Accuracy |
| --- | --- |
| Case 1 | 0.8229 |
| Case 2 | 0.8074 |
| Case 3 | 0.7946 |
| Case 4 | 0.7886 |
| **Case 5** | **0.8667** |

A possible explanation for those results is that pretrained models perform better on larger texts and need every token that they might learn from. Even though stop words might not be used as markers in machine learning methods [28], the BERT model performed better when stop words were not removed. As mentioned, transfer learning techniques are features-independent, and their capability in contextually model language makes them powerful enough to understand natural language. Our goal was to test aspects that can affect the performance of such pretrained models, namely the most commonly used preprocessing techniques for profiling the age and gender of authors. As a result, our study offers valuable findings that shed light on the best preprocessing techniques that can be applied when using a pretrained model to profile authors by gender.

## 5. CONCLUSIONS

Our experiments on how preprocessing techniques impact the gender profiling of authors when using a transfer learning model confirmed that the BERT performs best when no preprocessing techniques are applied. They also revealed that removing stop words lowers the accuracy by 1%. Those results indicate that pretrained models perform better when longer texts are present. Other common preprocessing techniques in the literature were included in the experiments and showed that they affect the pretrained model performance negatively. On top of that, our findings suggest that the use of pretrained models could be standardized, for it does not rely on many dependent parameters such as preprocessing and variable features.

For the future, the study has provided the groundwork for using transfer learning techniques to advance the field of author profiling, with findings that can serve as a starting point for using transfer learning in author profiling. Although the scope of the study was limited in terms of the pretrained model and the number of preprocessing techniques used, future work could involve using more than one pretrained model in a bid to better generalize the findings. Researchers could also apply more preprocessing techniques to cover more preprocessing possibilities.

**REFERENCES**

[1] S. Keretna, A. Hossny, and D. Creighton, "Recognising user identity in twitter social networks via text mining," Proc. - 2013 IEEE Int. Conf. Syst. Man, Cybern. SMC 2013, pp. 3079–3082, 2013, doi: 10.1109/SMC.2013.525.

[2] T. Yepes, Ray (ATX Forensics LLC, Austin, "The Art of Profiling in a Digital World.pdf." International Association of Chiefs of Police, p. 8, 2016.

[3] A. Halimi and E. Ayday, "Profile Matching Across Unstructured Online Social Networks: Threats and Countermeasures *."

[4] A. Rocha et al., "Authorship Attribution for Social Media Forensics," IEEE Trans. Inf. Forensics Secur., vol. 12, no. 1, pp. 5–33, 2017, doi: 10.1109/TIFS.2016.2603960.

[5] S. Argamon, M. Koppel, J. W. Pennebaker, and J. Schler, "Automatically profiling the author of an anonymous text," Commun. ACM, vol. 52, no. 2, pp. 119–123, 2009, doi: 10.1145/1461928.1461959.

[6] M. De-Arteaga, S. Jimenez, G. Dueñas, S. Mancera, and J. Baquero, "Author profiling using corpus statistics, lexicons and stylistic features: Notebook for PAN at CLEF-2013," CEUR Workshop Proc., vol. 1179, 2013.

[7] R. Bayot and T. Goncalves, "Multilingual author profiling using word embedding averages and SVMs," Ski. 2016 - 2016 10th Int. Conf. Software, Knowledge, Inf. Manag. Appl., pp. 382–386, 2017, doi: 10.1109/SKIMA.2016.7916251.

[8] M. Agrawal and T. Gonçalves, "Age and Gender Identification using Stacking for Classification★ Notebook for PAN at CLEF 2016," in CEUR Workshop Proceedings, 2016, vol. 18, no. 24, p. 28.

[9] L. Miculicich Werlen, "Statistical Learning Methods for Profiling Analysis Notebook for PAN at CLEF 2015," CLEF 2015 Labs Work. Noteb. Pap., 2015.

[10] A. Vaswani et al., "Attention is all you need," Adv. Neural Inf. Process. Syst., vol. 2017-Decem, no. Nips, pp. 5999–6009, 2017.

[11] J. Devlin, M. W. Chang, K. Lee, and K. Toutanova, "BERT: Pre-training of deep bidirectional transformers for language understanding," NAACL HLT 2019 - 2019 Conf. North Am. Chapter Assoc. Comput. Linguist. Hum. Lang. Technol. - Proc. Conf., vol. 1, no. Mlm, pp. 4171–4186, 2019.

[12] J. Howard and S. Ruder, "Universal language model fine-tuning for text classification," ACL 2018 - 56th Annu. Meet. Assoc. Comput. Linguist. Proc. Conf. (Long Pap., vol. 1, pp. 328–339, 2018, doi: 10.18653/v1/p18-1031.

[13] F. Rangel, P. Rosso, M. Koppel, E. Stamatatos, and G. Inches, "Overview of the author profiling task at PAN 2013," CEUR Workshop Proc., vol. 1179, pp. 8–11, 2013.

[14] F. Rangel and I. Chugur, "Overview of the 2nd Author Profiling Task at PAN 2014," 2014.

[15] F. Rangel, F. Celli, P. Rosso, M. Potthast, B. Stein, and W. Daelemans, "Overview of the 3rd Author Profiling Task at PAN 2015," in CEUR Workshop Proceedings, 2015, vol. 1179, [Online]. Available: http://pan.webis.de.

[16] F. Rangel, P. Rosso, B. Verhoeven, W. Daelemans, M. Potthast, and B. Stein, "Overview of the 4th author profiling task at PAN 2016: Cross-genre evaluations," CEUR Workshop Proc., vol. 1609, pp. 750–784, 2016.

[17] F. Rangel, P. Rosso, M. Potthast, and B. Stein, "Overview of the 5th Author Profiling Task at PAN 2017: Gender and Language Variety Identification in Twitter," CEUR Workshop Proc., vol. 2380, 2017.

[18] H. Gómez-Adorno, I. Markov, G. Sidorov, J. P. Posadas-Durán, M. A. Sanchez-Perez, and L. Chanona-Hernandez, "Improving Feature Representation Based on a Neural Network for Author Profiling in Social Media Texts," Comput. Intell. Neurosci., vol. 2016, 2016, doi: 10.1155/2016/1638936.

[19] E. Lundeqvist and M. Svensson, "Author profiling: A machine learning approach towards detecting gender, age, and native language of users in social media," no. 17013, p. 81, 2017.

[20] S. Mamgain, R. C. Balabantaray, and A. K. Das, "Author profiling: Prediction of gender and language variety from document," Proc. - 2019 Int. Conf. Inf. Technol. ICIT 2019, pp. 473–477, 2019, doi: 10.1109/ICIT48102.2019.00089.

[21] R. Bakkar Deyab, J. Duarte, and T. Gonçalves, "Author Profiling Using Support Vector Machines Notebook for PAN at CLEF 2016," in CEUR Workshop Proceedings, 2016, pp. 2–5.

[22] G. Kheng, L. Laporte, and M. Granitzer, "INSA Lyon and UNI passau's participation at PAN@CLEF'17: Author Profiling task: Notebook for PAN at CLEF 2017," in CEUR Workshop Proceedings, 2017, vol. 1866.

[23] M. Martinc, I. Skrjanec, K. Zupan, and S. Pollak, "PAN 2017: Author Profiling-Gender and Language Variety Prediction.," in Working Notes of CLEF 2017-Conference and Labs of the Evaluation Forum, Ireland, 11-14 September, 2017.

[24] S. S. R. Seelam, S. Kumar, C. M. Gopi, and R. T. Raghunadha, "A New Term Weight Measure for Gender and Age Prediction of the Authors by analyzing their Written Texts," Proc. 8th Int. Adv. Comput. Conf. IACC 2018, pp. 150–156, 2018, doi: 10.1109/IADCC.2018.8692092.

[25] F. Rangel, P. Rosso, M. Koppel, E. Stamatatos, and G. Inches, "Overview of the author profiling task at PAN 2013," CEUR Workshop Proc., vol. 1179, 2013.

[26] S. Panigrahi, A. Nanda, and T. Swarnkar, "A Survey on Transfer Learning," Smart Innov. Syst. Technol., vol. 194, pp. 781–789, 2021, doi: 10.1007/978-981-15-5971-6_83.

[27] Q. Yang, Y. Zhang, W. Dai, and S. J. Pan, "Transfer Learning in Natural Language Processing," Transf. Learn., pp. 234–256, 2020, doi: 10.1017/9781139061773.020.

[28] T. R. Reddy, B. V. Vardhan, and P. V. Reddy, "N-gram approach for gender prediction," Proc. - 7th IEEE Int. Adv. Comput. Conf. IACC 2017, pp. 860–865, 2017, doi: 10.1109/IACC.2017.0176.